\documentclass[letterpaper]{article} % DO NOT CHANGE THIS
\usepackage{aaai2026}  % DO NOT CHANGE THIS
\usepackage{times}  % DO NOT CHANGE THIS
\usepackage{helvet}  % DO NOT CHANGE THIS
\usepackage{courier}  % DO NOT CHANGE THIS
\usepackage[hyphens]{url}  % DO NOT CHANGE THIS
\usepackage{graphicx} % DO NOT CHANGE THIS
\usepackage{natbib}  % DO NOT CHANGE THIS AND DO NOT ADD ANY OPTIONS TO IT
\usepackage{caption} % DO NOT CHANGE THIS AND DO NOT ADD ANY OPTIONS TO IT
\usepackage{algorithm}
\usepackage{algorithmic}

\newcommand{\sysname}{DriveLiDAR4D}
\newcommand{\modelname}{LiDAR4DNet}

\usepackage{amssymb}
\usepackage{amsmath}
\usepackage{colortbl}
\usepackage[table]{xcolor}

\definecolor{color_ours}{RGB}{229,240,219}  
\usepackage{siunitx}
\newcommand{\mypara}[1]{\vspace{1mm}\noindent\textbf{#1}}

\usepackage{multirow}
\usepackage{booktabs}

\usepackage{nicefrac}
\usepackage{comment}

\usepackage{cleveref}
\crefname{figure}{Fig.}{Figs.}
\crefformat{equation}{(#2#1#3)}
\crefrangeformat{equation}{(#3#1#4) to (#5#2#6)}
\crefmultiformat{equation}{(#2#1#3)}{和(#2#1#3)}{，(#2#1#3)}{和(#2#1#3)}
\crefname{section}{Sec.}{Secs.}
\Crefname{section}{Sec.}{Secs.}
\crefname{table}{Tab.}{Tabs.}
\Crefname{table}{Tab.}{Tabs.}

\definecolor{iccvblue}{rgb}{0.21,0.49,0.74}
\usepackage{newfloat}
\usepackage{listings}
\DeclareCaptionStyle{ruled}{labelfont=normalfont,labelsep=colon,strut=off} % DO NOT CHANGE THIS
\floatstyle{ruled}
\newfloat{listing}{tb}{lst}{}
\floatname{listing}{Listing}
\title{\sysname: Sequential and Controllable LiDAR Scene Generation \\for Autonomous Driving}
\author{
    Kaiwen Cai\textsuperscript{\rm 1}\equalcontrib,
    Xinze Liu\textsuperscript{\rm 1}\equalcontrib,
    Xia Zhou\textsuperscript{\rm 1}\equalcontrib,
    Hengtong Hu\textsuperscript{\rm 1},
    Jie Xiang\textsuperscript{\rm 1}, \\
    Luyao Zhang\textsuperscript{\rm 1},
    Xueyang Zhang\textsuperscript{\rm 1},
    Kun Zhan\textsuperscript{\rm 1},
    Yifei Zhan\textsuperscript{\rm 1},
    Xianpeng Lang\textsuperscript{\rm 1}
}
\affiliations{
    \textsuperscript{\rm 1}Li Auto Inc.\\
    caikaiwen1@lixiang.com
}

\usepackage{bibentry}
\begin{document}

\maketitle

\begin{abstract}
The generation of realistic LiDAR point clouds plays a crucial role in the development and evaluation of autonomous driving systems. Although recent methods for 3D LiDAR point cloud generation have shown significant improvements, they still face notable limitations, including the lack of sequential generation capabilities and the inability to produce accurately positioned foreground objects and realistic backgrounds. These shortcomings hinder their practical applicability. In this paper, we introduce \sysname, a novel LiDAR generation pipeline consisting of multimodal conditions and a novel sequential noise prediction model \modelname, capable of producing temporally consistent LiDAR scenes with highly controllable foreground objects and realistic backgrounds. To the best of our knowledge, this is the first work to address the sequential generation of LiDAR scenes with full scene manipulation capability in an end-to-end manner. We evaluated \sysname\ on the nuScenes and KITTI datasets, where we achieved an FRD score of 743.13 and an FVD score of 16.96 on the nuScenes dataset, surpassing the current state-of-the-art (SOTA) method, UniScene, with an performance boost of  37.2\% in FRD and 24.1\% in FVD, respectively.

\end{abstract}

% temporal nuscenes rebuttal update
% FRD: ours=743.13 vs. uniscene=1182.94
% (1182.94-743.13)/1182.94=37.2%
% FVD: ours=16.96 vs. uniscene=21.04
% (21.04-16.96)/16.96=24.1%
\section{Introduction}

\begin{figure}[ht]
    \begin{center}
  \includegraphics[width=1.0\linewidth]{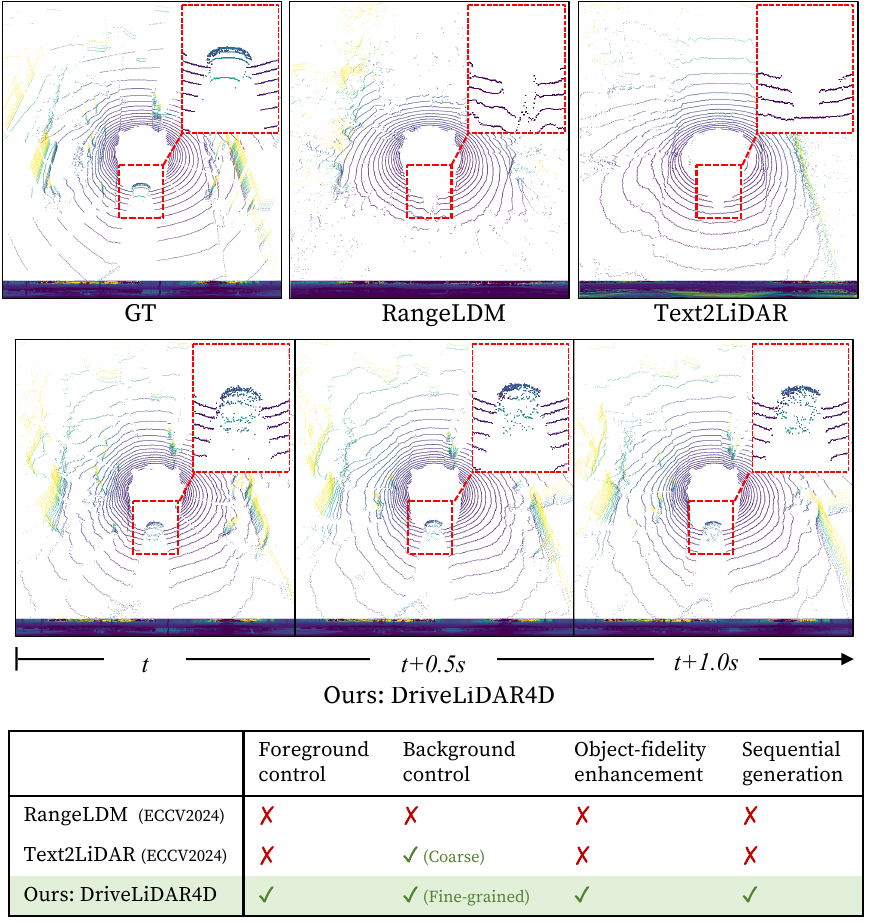}
    \end{center}
       \caption{Comparison of  LiDAR scenes generated by different methods on the nuScenes val split. \sysname\ is the first work to achieve sequential LiDAR scene generation with highly controllable scene manipulation abilities, including foreground control, background control and object-fidelity enhancement.}
\label{fig:fig_openfig}
 \end{figure}

Data is a foundational element driving artificial intelligence advances.  Within autonomous driving research, high-quality data is particularly crucial due to: i) the inherent data-intensive requirements of deep learning models, and ii) the necessity of capturing corner cases — rare driving behaviours and uncommon road environments — which are essential for developing safety-critical systems. However, collecting and annotating diverse multi-modal datasets (e.g., camera and LiDAR) remains time-consuming and resource-intensive. While recent generative models have demonstrated promising capabilities for synthesizing visual data, LiDAR scene generation—despite its critical role in providing geometric awareness — remains comparatively underdeveloped.  In this work, we aim to advance existing LiDAR scene generation techniques to better address real-world autonomous driving requirements.

% objects control - road sketch control
To synthesize realistic LiDAR data that accurately captures diverse real-world traffic scenarios, a LiDAR scene generation method should suppport flexible customization of road layouts and dynamic object placements. Recent studies, such as LiDARGen~\cite{zyrianov2022learning}, UltraLiDAR~\cite{xiong2023ultralidar}, R2DM~\cite{nakashima2024lidar}, and RangeLDM~\cite{hu2025rangeldm}, have made considerable advancements in producing realistic LiDAR data.  However, these techniques predominantly generate data in an unconditional manner, lacking the capability to manipulate specific scene elements.

% background control - scene caption control
The quality of background LiDAR point clouds is as critical as that of foreground objects to ensure realism.
To this end, Text2LiDAR~\cite{wu2024text2lidar} utilizes textual descriptions as conditioning input. Nonetheless, this method is confined to coarse descriptions encompassing weather conditions, time of day, and object names. The omission of detailed background information, such as specific descriptions of trees or buildings, compromises the realism of the generated LiDAR data. 

% sequential scene generation
In addition, current 3D LiDAR scene generation methods prove insufficient when it comes to capturing the dynamic behaviours of objects. To mitigate this issue, LidarDM \cite{zyrianov2024lidardm} proposes to generate LiDAR sequences by separately modeling static scenes and dynamic objects. However, it lacks background control, and its two-stage compositional strategy potentially compromises the realism and coherence of the resulting point cloud distributions.

In summary, current LiDAR scene generation methods are notably deficient when it comes to integrating all critical capbalities: i) sequential scene generation with detailed control over both ii) foreground and iii) background components.  
To bridge this gap, we propose \sysname, an end-to-end 4D LiDAR scene generation pipeline that facilitates the generation of sequential LiDAR scenes with comprehensive scene manipulation capabilities. 
\sysname\ is distinguished by two principal features: i) the incorporation of multimodal conditions, including scene captions, road sketches and object priors, and ii) a meticulously designed equirectangular spatial-temporal noise prediction model, \modelname, which ensures spatial and temporal consistency throughout denoising processes. 

\cref{fig:fig_openfig} displays LiDAR scenes generated by different methods on the nuScenes val split. It is evident that RangeLDM \cite{hu2025rangeldm} and Text2LiDAR \cite{wu2024text2lidar} are unable to generate the vehicle accurately, and backgrounds do not align with those observed in the Ground Truth(GT) scene. In contrast, the vehicles' positions and structures, and the background generated by \sysname\ closely align with those in the GT scene. Furthermore, it is noteworthy that \sysname\ is capable of generating a sequence of LiDAR scenes that maintain temporal consistency, whereas RangeLDM \cite{hu2025rangeldm} and Text2LiDAR \cite{wu2024text2lidar} can only create individual LiDAR scenes in isolation.

To summarize, our specific contributions are as follows:
\begin{itemize}
    \item This is the first work to achieve precise control over foreground objects, including manipulation of their positions and sizes, and fine-grained control over background elements in LiDAR scene generation.  
    \item We propose a novel equirectangular spatio-temporal diffusion model, \modelname,  which achieves end-to-end \textit{sequential} LiDAR scene generation while ensuring consistency over foreground and background elements.
    \item We demonstrate the effectiveness of our proposed \sysname\ on the KITTI and nuScenes datasets, where it outperforms the current SOTA methods.
\end{itemize}

\section{Related Work}
\subsection{Diffusion Models}
Diffusion models have been the de-facto choice in image and video generation tasks \cite{podell2023sdxl, gupta2024photorealistic, yang2024cogvideox}. 

Research in this field encompasses various aspects of improvement in     diffusion policy \cite{croitoru2023diffusion, liu2023flow}, model architecture \cite{bar2024lumiere,peebles2023scalable}, conditioning strategies \cite{zhang2023adding, khachatryan2023text2video}, among others.
For a comprehensive survey of image and video generation, we refer readers to~\cite{cao2024survey}. 
However, these methods are primarily designed for visual modality data. LiDAR data presents distinct challenges compared to visual data, as it describes a scene with unordered and unevenly distributed points. This fundamental difference makes it challenging to effectively leverage diffusion models for the LiDAR modality.

\subsection{LiDAR Scene Generation}

Recent LiDAR data generation approaches can be categorized as either unconditional or conditional generation. In the unconditional category, methods such as LiDARGen \cite{zyrianov2022learning}, R2DM \cite{nakashima2024lidar}, and LidarGRIT \cite{haghighi2024taming} train diffusion models directly on the pixel space, while RangeLDM additionally employs VAE \cite{kingma2013auto} to compress equirectangular images. In the conditional category, Text2LiDAR \cite{wu2024text2lidar} conditions the diffusion process on coarse textual scene descriptions. LiDM \cite{ran2024towards} investigates various conditioning inputs but applies each condition independently, limiting its comprehensive manipulation capabilities. Our object priors condition shares similarities with OLiDM \cite{yan2024olidm}, which utilizes synthetic objects as conditions for scene generation. However, OLiDM \cite{yan2024olidm} is designed to generate single LiDAR scenes, whereas our approach incorporates multiple multimodal conditions to generate \textit{sequential} LiDAR scenes—a non-trivial task requiring careful pipeline design. LidarDM \cite{zyrianov2024lidardm} decomposes the generation task into two stages, static and dynamic object synthesis. But this approach compromises the realism of object point clouds due to its simplistic modeling of 3D object structures. Moreover,

We also see a growing research on jointly generating LiDAR point clouds and images. XDrive~\cite{xie2025xdrive} uses two diffusion models to generate image and LiDAR point clouds simultaneously, where latent LiDAR features and latent image features attend to each other via epipolar guidance. Albeit, XDrive \cite{xie2025xdrive} is limited to single LiDAR scene generation.
UniScene~\cite{li2024uniscene} indirectly enforces temporal consistency across sequential frames by mediating cross-modal interactions through implicit conditioning on 3D occupancy priors. Nevertheless, it exhibits limitations in synthesizing high-fidelity objects.

In summary, current methodologies in LiDAR generation field predominantly focus on 3D scene generation with coarse conditioning, neglecting the integration of complete temporal coherence with fine-grained controllability. To the best of our knowledge,  \sysname\ is the first unified framework to resolve the aforementioned limitations by integrating a equirectangular spatio-temporal diffusion model with multimodal conditioning capabilities.

\section{\sysname}

\label{sec_seqlidar}

\cref{fig_diagram_seqnet} illustrates the pipeline of \sysname: During training, we first derive the three multimodal conditions: road sketches, scene captions and object priors. Then, \modelname\ takes as input a sequence of noised equirectangular images, conditioned on the three multimodal conditions, and predicts the added noises. During inference, \modelname\ reconstructs the sequence of equirectangular images, again utilizing the three multimodal conditions.
In this section, we first detail the multimodal conditions and then elaborate on the specifics of \modelname.

\subsection{Preliminary}
Our method is based on diffusion models, and we briefly introduce its principle in this section. We employ the denoising diffusion probabilistic model (DDPM). In DDPM, a forward diffusion process will gradually destroy a sample $\boldsymbol{x}$ by adding Gaussian noise at each timestep $t$: $\boldsymbol{x}_t=\alpha_t \boldsymbol{x} + \sigma_t \epsilon$. The variable $\boldsymbol{x}_t$ will finally arrive at Gaussian distribution, which can be written by
\begin{equation}
    q(\boldsymbol{x}_t|\boldsymbol{x}) = \mathcal{N} (\alpha_t\boldsymbol{x}, \sigma^2_t \boldsymbol{I}), 
\end{equation}
where $\alpha_t$ and $\sigma_t$ are hyperparameters that determine the noise scheduling. We follow the $\alpha$-cosine schedule \cite{saharia2022photorealistic} where $\alpha_t=\cos(\pi t/2)$, $\sigma_t=\sin(\pi t/2)$. Consequently, the intermediate transition process of variable from timestep $s$ to $t$ ($0<s<t<1$) can be formulated as 
\begin{equation}
    q(\boldsymbol{x}_t|\boldsymbol{x}_s)= \mathcal{N} (\alpha_{t|s}\boldsymbol{x}_s, \sigma^2_{t|s} \boldsymbol{I}), 
\end{equation}
where $\alpha_{t|s}={\alpha_t}/{\alpha_s}$, $\sigma^2_{t|s}=\sigma^2_t - \alpha^2_{t|s}\sigma^2_s$.

A reverse diffusion process is used to denoise the sample in the latent space, which can be represented as
\begin{equation}
\label{eq_p_step}
    p(\boldsymbol{x}_s|\boldsymbol{x}_t) = \mathcal{N}(\boldsymbol{\mu}_t(\boldsymbol{x}, \boldsymbol{x}_t), \Sigma^2_t\boldsymbol{I}),
\end{equation}
where $\boldsymbol{\mu}_t(\boldsymbol{x}, \boldsymbol{x}_t)=(\alpha_{t|s}\sigma^2_s/\sigma^2_t) \cdot \boldsymbol{x}_t + (\alpha_s\sigma^2_{t|s}/\sigma^2_t) \cdot \boldsymbol{x}$, $\Sigma^2_t=\sigma^2_{t|s}\sigma^2_s/\sigma^2_t$.

During training, with a noised sample $\boldsymbol{x}_t$, the model is learning to estimate the unknown $\boldsymbol{x}_s$ via  estimating $\boldsymbol{x}$ in \cref{eq_p_step}. Practically, we adopt $\epsilon$-prediction \cite{saharia2022photorealistic} by reparameterizing $\boldsymbol{x}$ as a function of $\boldsymbol{x}_t$ and $\epsilon$. The loss function is denoted as 
\begin{equation}
    \mathcal{L} = \mathbb{E}_{\boldsymbol{x}, \boldsymbol{\epsilon} \sim \mathcal{N}(\mathbf{0}, \mathbf{I}, t)} \left [ \| \boldsymbol{\epsilon} - \hat{\boldsymbol{\epsilon}}(\boldsymbol{x}_t, t, \boldsymbol{c}) \| ^2 \right ],
\end{equation}
where $\boldsymbol{c}$ means conditions (will be described in \cref{sec_seqlidar}).

Once the model is trained, we evaluate \cref{eq_p_step} iteratively as $t: 1 \rightarrow 0$ and the final sample is expected to approximate the data distribution. We set the number of iterations as 256 in denoising process.

\subsection{Multimodal Conditions}
\label{sec_conditions}

\begin{figure}[t]
    \begin{center}
       \includegraphics[width=\linewidth]{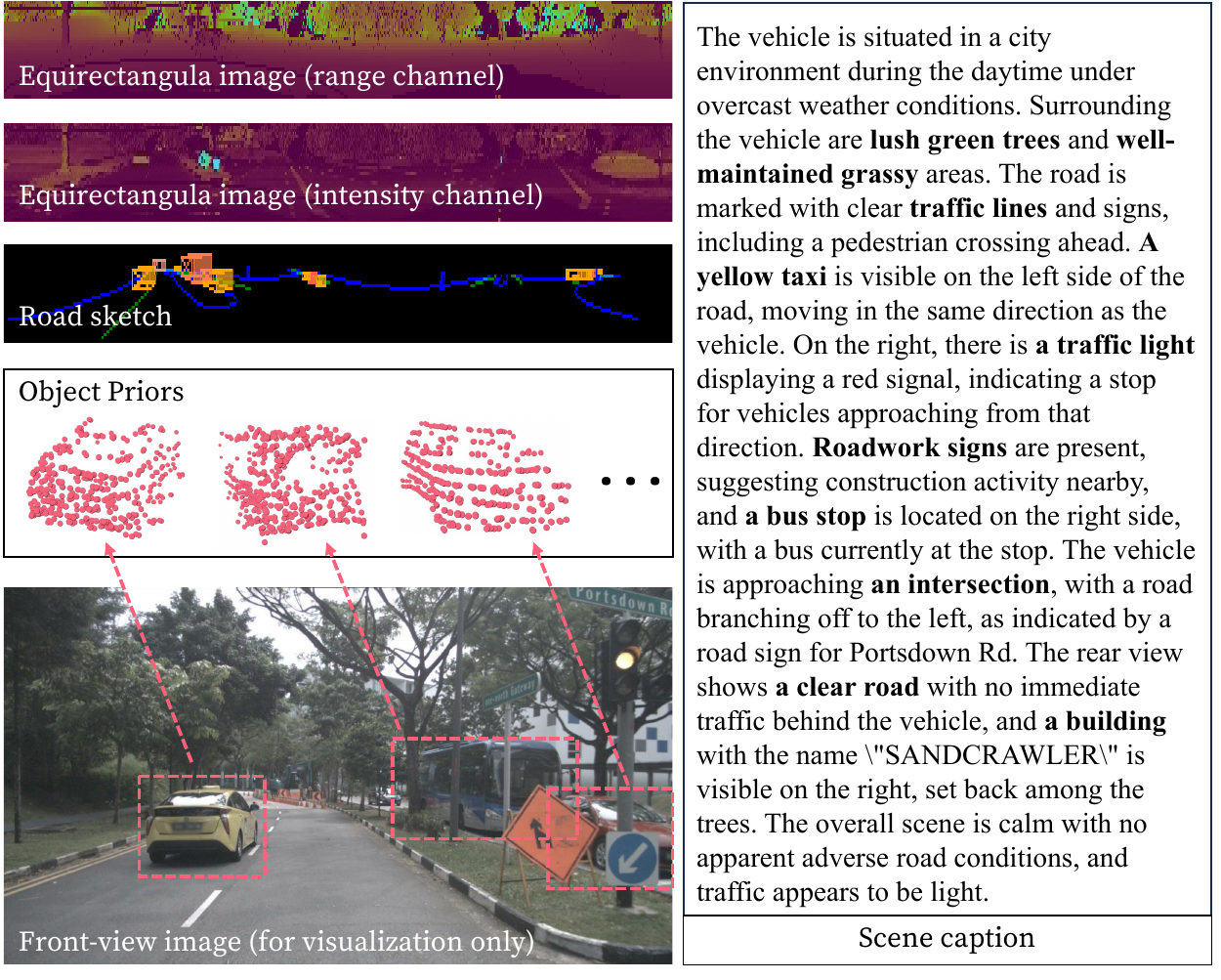}
    \end{center}
       \caption{Visualization of the multimodal conditions of an example from the nuScenes dataset (Images have been resized for better visualization). }
       \label{fig_condition}
 \end{figure}

\noindent \textbf{Road Sketch}:
The road sketch incorporates road layouts and object-specific information.
Firstly, road layouts are delineated through curbs and lane lines. This representation provides a pixel-to-pixel control of the spatial organization of the road.
Secondly, object-specific information is instantiated as 3D bounding boxes, which encapsulate the size, location, and heading direction of each object. 
\begin{comment}
This representation offers the following advantages:
1) It facilitates precise object-level manipulations, allowing for detailed control over individual elements within the scene. And 2) The use of multiple 3D bounding boxes implicitly conveys occlusion states, providing depth and spatial relationship information without explicit encoding.
\end{comment}

To integrate the two components into a unified representation, both the road layouts and 3D bounding boxes are projected onto the equirectangular image (will be detailed soon) plane of the LiDAR sensor. This results in a road sketch that has a same size as the input equirectangular image.  \cref{fig_condition} depicts an example of road sketch condition from the nuScenes dataset.

 \begin{figure*}[t]
    \begin{center}
       \includegraphics[width=1\linewidth]{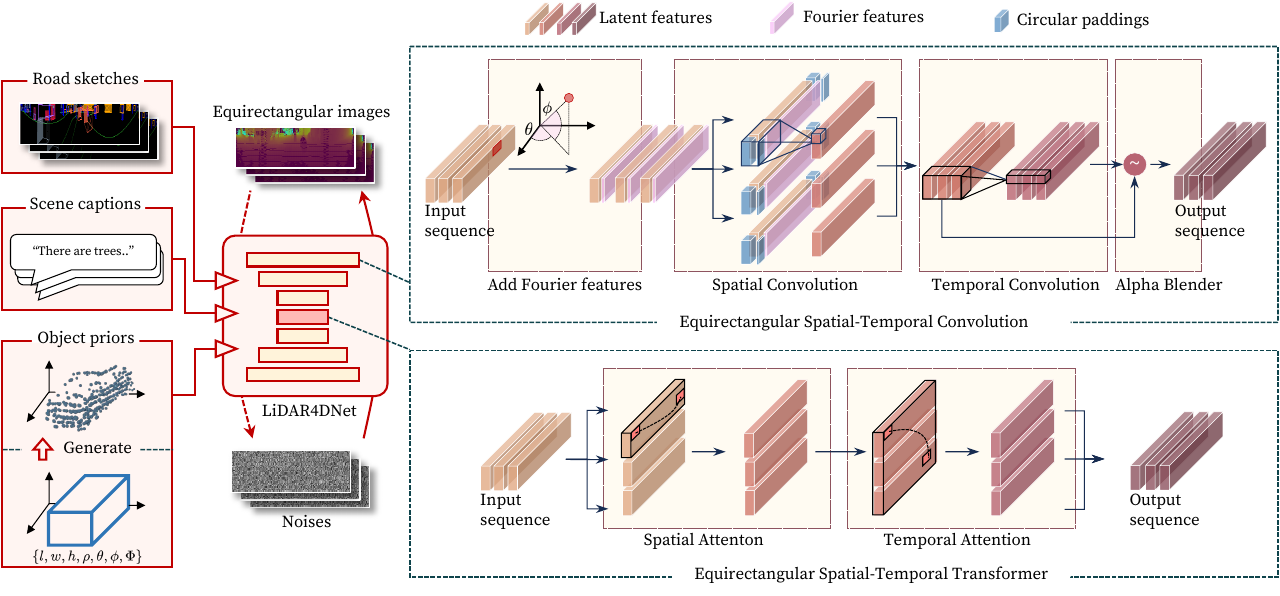}
    \end{center}
       \caption{The pipeline of \sysname. We first derive multi-modal conditions, including road sketches, scene captions and object priors from a given road scene (see \cref{sec_conditions}).  Then, the proposed \modelname\ predicts the sequential noises based on the multimodal conditions, where Equirectangular Spatial-Temporal Convolution (EST-Conv) and Equirectangular Spatial-Temporal Convolution (EST-Trans) enforce spatial and temporal consistency (see \cref{sec_seqnet} ).}
       \label{fig_diagram_seqnet}
 \end{figure*}

\noindent \textbf{Scene Caption}:
In addition to foreground control by road sketches, we propose to use scene captions that provide a comprehensive description of the background. Unfortunately, existing LiDAR datasets lack high-quality scene captions: the KITTI-360 dataset \cite{Liao2022PAMI} does not include scene captions, while the nuScenes dataset \cite{caesar2020nuscenes} provides only brief, one-sentence descriptions of the weather or time of day. 
Therefore, we leverage capabilities of the powerful vision-language model GPT-4V \cite{hurst2024gpt} to generate detailed scene captions. Specifically, we input the surrounding images into GPT-4V and request descriptive text for the depicted scene. 
\cref{fig_condition} depicts an example of the scene caption condition from the nuScenes  dataset. In this example, the scene caption encompasses not only the foreground object "taxi", but also the background elements such as "trees" and "grassy areas".

\noindent \textbf{Object Priors}:
In autonomous driving scenarios, LiDAR measurements of different elements are often imbalanced; for example, a vehicle may be represented by hundreds of points, whereas the ground may encompass thousands. Diffusion models typically attempt to approximate the entire scene without addressing this imbalance, which can result in suboptimal generation quality for objects.
To mitigate this issue, we propose initially synthesizing the point clouds of objects, which then serve as a condition to guide the model in generating the entire scene. The underlying intuition is that while road sketches indicate the locations of objects, the distribution of points for those objects is not adequately guided. Incorporating synthetic object point clouds as priors provides a stronger guidance during the denosing process.

We train an object generation model, DiT-3D \cite{mo2023ditd}, conditioned on the object's category, size, polar coordinates relative to the LiDAR sensor, and heading direction. We denote these conditions as \(\{category, l, w, h, \rho, \theta, \phi, \Phi\}\). The object priors are obtained by generating the point clouds of the objects using the pretrained object generation model. We then project these object points onto the equirectangular image plane of the LiDAR sensor.

\subsection{\modelname}

\label{sec_seqnet}

We follow \cite{nakashima2024lidar, wu2024text2lidar} and adopt equirectangular representation due to its efficiency in describing large scenes.  A LiDAR sensor with $H \times W$ spatial resolution captures range and reflectance measurements at $W$ azimuth angles and $H$ elevation angles, resulting in an equirectangular image $\boldsymbol{x} \in \mathbb{R}^{2\times H \times W}$. 
%Similar to \cite{nakashima2024lidar}, 
We scale a range variable using ${\hat{d}}= \nicefrac{\log{(d+1)}}{\log{(d_{max}+1)}}$ and subsequently normalize it to the range $[-1, 1]$.

As is depicted in \cref{fig_diagram_seqnet}, the proposed \modelname\ is a UNet-like encoder-decoder model \cite{saharia2022photorealistic}, incorporating stacked Equirectangular Spatial-Temporal Convolution (EST-Conv) modules across four scales, along with an Equirectangular Spatial-Temporal Transformer (EST-Trans) module at the bottleneck.

\noindent \textbf{EST-Conv}:
Research in RGB image generation has explored 3D convolutions for learning temporal features. However, equirectangular images possess distinct characteristics compared to standard RGB images:  1) the pixel coordinates $(\theta, \phi)$ exhibit strong correlation with LiDAR data patterns (e.g., regions that are in close proximity and are positioned low tend to represent the ground), 2) the left and right boundaries of equirectangular images represent a continuous region in LiDAR measuring space (particularly for LiDAR sensors with 360-degree horizontal field-of-view (FOV)), and 3) the pixel value distribution differs significantly from that of RGB images. The unique properties of equirectangular images make the optimum solution for processing sequential equirectangular images unknown.

\begin{table*}[t]
\centering
\begin{tabular}{l|c|ccc|ccc} 
\toprule
                                      & \multirow{2}{*}{Venues} & \multicolumn{3}{c|}{KITTI-360}     & \multicolumn{3}{c}{nuScenes}                                       \\
                                      &                         & FRD$\downarrow$                       & MMD$^1$$\downarrow$  & JSD$\downarrow$            & FRD$\downarrow$     & MMD$^1$$\downarrow$  & JSD$\downarrow$             \\ 
\midrule
LiDARGen \cite{zyrianov2022learning} & ECCV2022                & 2040.1             & 3.87          & 0.067          & -                 & 19.00         & 0.160           \\
R2DM$^2$ \cite{nakashima2024lidar}        & ICRA2024                & \underline{275.67}  & 5.55          & 0.049          & -                 & -             & -               \\
Text2LiDAR$^2$ \cite{wu2024text2lidar}    & ECCV2024                & 831.64             & 4.36          & 0.044          & -                    & -             & -               \\
RangeLDM$^2$ \cite{hu2025rangeldm}        & ECCV2024                & 2022.71             & \textbf{0.75} & \textbf{0.035} & \underline{492.49}  & \underline{2.75}  & \underline{0.054}   \\
%LidarDM \cite{zyrianov2024lidardm}        & ICRA2025                & -             & \underline{1.67} & 0.119 & -  & -  & -   \\
\rowcolor{color_ours}
\sysname               &                         & \textbf{244.25}  & \underline{3.31}  & \underline{0.042}  & \textbf{210.55} & \textbf{1.84} & \textbf{0.045}  \\
\bottomrule
\multicolumn{5}{l}{\footnotesize 1: MMD has been multiplied by \({10^{4}}.\)} \\
\multicolumn{5}{l}{\footnotesize 2: We reproduced the results with their released source codes.} \\
\end{tabular}
\centering
\caption{Unconditional generation results of different methods on the KITTI-360 test set and the nuScenes val split.}
\label{tabel1}

\end{table*}

Therefore, we propose EST-Conv, a novel approach designed to learn spatial and temporal features from equirectangular images.
In EST-Conv, to enhance spatial consistency, we leverage the correlation between pixel coordinates and LiDAR measurements by extracting Fourier features \cite{tancik2020fourier} for each pixel coordinate $(\theta, \phi)$ and concatenating these features with the input equirectangular images. The integration of Fourier features helps the model better capture the underlying geometric patterns in LiDAR data.
Moreover, equirectangular images capture a 360-degree horizontal field of view, implying that the left and right boundaries represent continuous 3D space. To better model this spatial continuity, we replace standard zero-padding in 2D convolutions with circular padding, enhancing feature learning across horizontal image boundaries. Sequential equirectangular features are expaned along the batch dimension prior to being processed by 2D convolutional operations.  

To enforce temporal consistency, we adopt 3D convolutions to directly process sequential equirectangular features. The unique geometry of equirectangular projections means adjacent pixels may represent distant objects in 3D space. Therefore, we implement 3D convolutions with small kernel sizes (d=3, h=1, w=1). Finally, we employ an Alpha-blender \cite{blattmann2023stable} to combine spatial and temporal equirectangular features. The Alpha-blender can be written as $\boldsymbol{y}=\alpha\cdot \boldsymbol{x}_S+(1-\alpha)\cdot\boldsymbol{x}_T$, where $\alpha$ is a learnable parameter, $\boldsymbol{x}_S$ and $\boldsymbol{x}_T$ are spatial and temporal features respectively.

\noindent \textbf{EST-Trans}:
Convolutions efficiently process features with progressively increasing receptive fields but exhibit limitations in capturing long-range correlations. On the other hand, attention mechanisms can effectively model long-range dependencies, albeit with significant computational overhead.
To effectively model long-range dependencies, we introduce EST-Trans and apply it exclusively at the bottleneck layer as illustrated in \cref{fig_diagram_seqnet}. We emperically found that this design strike a good balance between performance and efficiency.
In EST-Trans, input sequences are stacked along the batch dimension before applying spatial attention and then expanded before applying temporal attention. Spatial attention captures relationships across individual equirectangular images, while temporal attention facilitates feature interactions among sequential equirectangular images. This architecture minimizes computational costs while simultaneously improving both spatial and temporal consistency.

\noindent \textbf{Injecting Multimodal Conditions}: 
Due to the distinct characteristics of the three conditioning types, we implement tailored strategies for each. For road sketches, we employ channel concatenation due to their precise pixel-to-pixel correspondence with equirectangular images. For object priors, which may object involve shape transformations, we leverage ControlNet \cite{zhang2023adding} conditioning strategy to provide enhanced learning capacity. For scene captions, in addition to applying cross-attention \cite{saharia2022photorealistic}, we fuse timestep variables with captions to prevent captions’ influence from being overshadowed by the other two conditions. Additional architecture details are provided in the supplementary materials.

\section{Experimental Results}
\subsection{Experimental Settings}

\mypara{Datasets.}
We validated the effectiveness of \sysname\ on the two real-world autonomous driving datasets, the KITTI-360~\cite{Liao2022PAMI} and nuScenes \cite{nuscenes2019} datasets.
The KITTI-360 dataset features a Velodyne-64E LiDAR with 64 beams and 360-degree horizontal FOV. Following existing works~\cite{nakashima2024lidar, wu2024text2lidar}, we designated sequences 0 and 1 as the test set (26367 frames) and utilized the remaining sequences as training set (50348 frames). 
The nuScenes dataset provides 32-beam LiDAR measurements with 360-dergree horizontal FOV  at 20$\si{Hz}$ and keyframe annotations at 2$\si{Hz}$. 
Following common practice \cite{gao2023magicdrive}, we expanded the nuScenes bounding box labels to 12$\si{Hz}$ and adhered to the official training and validation dataset splits (700 training scenes and 150 validation scenes), resulting in 165280 training samples and 35364 validation samples.

\subsection{Unconditional LiDAR Scene Generation}
\label{sec_single_frame}

We compare \sysname\ against SOTA LiDAR scene generation methods,
LiDARGen~\cite{zyrianov2022learning}, R2DM~\cite{nakashima2024lidar}, Text2LiDAR\cite{wu2024text2lidar}, and RangeLDM~\cite{hu2025rangeldm}. 
\cref{tabel1} illustrates the unconditional generation results on the KITTI-360 test set and the nuScenes val split. 
It can be seen that \sysname\ achieves comparable generation quality than RangeLDM on the KITTI-360 dataset, with superior FRD \cite{wu2024text2lidar} and slightly inferior MMD \cite{nakashima2024lidar} and JSD \cite{nakashima2024lidar}. However, \sysname\ surpasses RangeLDM on the nuScenes dataset in all three metrics, FRD, MMD, and JSD. Overall, these results demonstrates superiority of \sysname\ among all compared methods.

\subsection{Conditional LiDAR Scene Generation}
\label{sec_conditional_generation}

\begin{table}[ht]
\centering
%\refstepcounter{table}
\scalebox{0.95}{
\begin{tabular}{lcc|ccc} 
\toprule
  & \begin{tabular}[c]{@{}c@{}}Road \\sketch\end{tabular} & \begin{tabular}[c]{@{}c@{}}Scene \\caption\end{tabular} & FRD$\downarrow$    & MMD$^1$$\downarrow$  & JSD$\downarrow$    \\ 
\midrule
Config A &                                                       &                                                         & 210.55 & 1.84 & 0.045  \\
Config B & \checkmark                                             &                                                         & 178.24 & 0.85 & 0.041  \\
Config C & \checkmark                                             & \checkmark                                               & \textbf{175.17} & \textbf{0.85} & \textbf{0.039}  \\
\bottomrule
\multicolumn{5}{l}{\footnotesize 1: MMD has been multiplied by \({10^{4}}.\)} \\
\end{tabular}}
\caption{
Conditional generation results of \sysname\ with different conditions on the nuScenes val split.}
\label{tab_ablation}
\end{table}

\begin{figure}[t]
    \begin{center}
       \includegraphics[width=\linewidth]{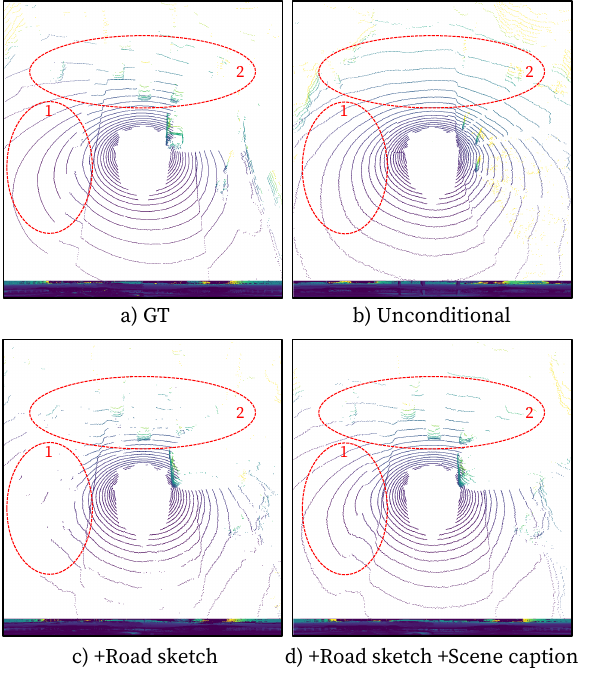}
    \end{center}
       \caption{LiDAR scenes generated by \sysname\ with different conditions on the nuScenes val split.}
       \label{fig_ablation}
\end{figure}

We examine the impacts of road sketch and scene caption conditions by training \modelname\ with subsets of the multimodal conditions.
\cref{tab_ablation} presents results of \sysname\ with different conditions on the nuScenes val split.
Obviously, the influence of road sketch conditioning is particularly significant, yielding a 15.34\% improvement in FRD (from 210.55 to 178.24) and a 53.80\% enhancement in MMD (from 1.84 to 0.85) compared to unconditional generation.
This is attributed to the precise information about the road structure and the foreground objects embedded within the road sketches. Scene captions contribute to further enhancing the realism of the generated point clouds by providing complementary contextual information.

\cref{fig_ablation} shows the qualitative comparison of the LiDAR scenes generated by \sysname\ with different conditions on the nuScenes val split. 
Without any conditioning, the generated LiDAR scene lacks specific object structures (as shown in \cref{fig_ablation}b). 
When conditioned solely on road sketches (as shown in \cref{fig_ablation}c), \sysname\ successfully controls object locations in the generated scene, though the background area (\textcolor{red}{red circle 1}) is incomplete. 
Upon incorporating scene caption conditioning (shown in \cref{fig_ablation}d), both foreground objects (\textcolor{red}{red circle 2}) and background elements (\textcolor{red}{red circle 1}) exhibit more complete structures, demonstrating the complementary benefits of this additional conditioning. This improvement stems from the comprehensive nature of scene captions, which encapsulate both foreground and background contextual information.

\begin{figure}[ht]
    \begin{center}
       \includegraphics[width=\linewidth]{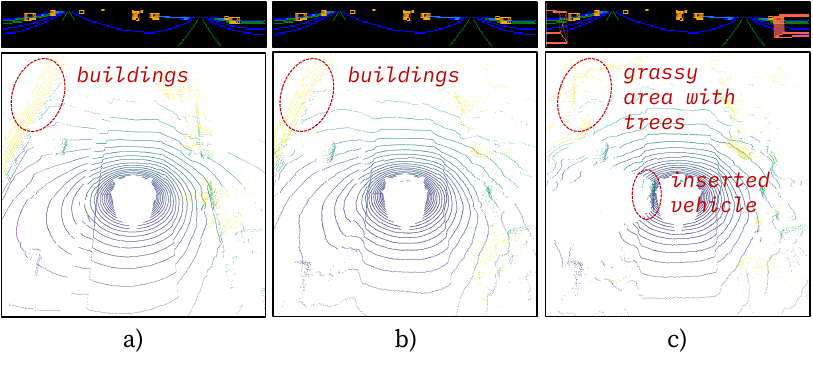}
    \end{center}
       \caption{Visualization of fine-grained scene manipulation of \sysname\ on the nuScenes val split. Top row: road sketches. Bottom row: LiDAR scenes. a) GT scene, b) Generated scene with multimodal conditions, c) Generated scene with \textit{edited} multimodal conditions.}
       \label{fig_fine_grained_control}
\end{figure}

Furthermore, \cref{fig_fine_grained_control} demonstrates an example of fine-grained manipulation of road elements in a generated LiDAR scene. In \cref{fig_fine_grained_control}b, we re-simulated the real scene with same multimodal conditions as those of GT scene, where we can see that the generated scene closely align with GT scene (e.g., see buildings). In \cref{fig_fine_grained_control}c, we inserted an vehicle and changed the background. This manipulation is achieved through edited multimodal conditions comprising 1) addition of a 3D box to the road sketch, 2) extraction of he corresponding object prior, and 3) modification of the scene caption from \{\textit{... building ...}\} to \{\textit{... grassy area with trees...}\}. The resulting scene successfully incorporates these manipulations while maintaining realism.

\subsection{Sequential and Controllable LiDAR Scene Generation}

\iffalse
\begin{table}[ht]
\scalebox{1}{
\begin{tabular}{l|ccccc} 
\toprule
         &  MMD$^1$$\downarrow$ & JSD$\downarrow$ & FRD$\downarrow$  & FVD$\downarrow$    \\ 
\midrule
LidarDM\textbf{}      & 25.38 &  0.153  & 1604.10  & 28.41  \\
UniScene\textbf{}      & \underline{21.22} &  \underline{0.141}  & \underline{1007.89}  & \underline{20.28}  \\
\rowcolor{color_ours}
\sysname   &  \textbf{2.13}    &  \textbf{0.049}  & \textbf{292.28}    & \textbf{14.86}  \\
\bottomrule
\multicolumn{5}{l}{\footnotesize 1: MMD has been multiplied by \({10^{4}}.\)} \\
\end{tabular}
}
\centering
\caption{Sequential generation results of different methods on the nuScenes val split.}
\label{tab_seq_con}

\end{table}
\fi

\begin{table}[ht]
\scalebox{1}{
\begin{tabular}{l|ccccc} 
\toprule
         &  MMD$^1$$\downarrow$ & JSD$\downarrow$ & FRD$\downarrow$  & FVD$\downarrow$    \\ 
\midrule
LidarDM\textbf{}      & 25.53 &  0.155  & 1800.18  & 28.48  \\
UniScene\textbf{}      & {21.66} &  {0.143}  & {1182.94}  & {21.04}  \\
\rowcolor{color_ours}
\sysname   &  \textbf{2.94}    &  \textbf{0.069}  & \textbf{743.13}    & \textbf{16.96}  \\

\bottomrule
\multicolumn{5}{l}{\footnotesize 1: MMD has been multiplied by \({10^{4}}.\)} \\
\end{tabular}
}
\centering
\caption{Sequential generation results of different methods on the nuScenes val split.}
\label{tab_seq_con}
\end{table}

\begin{figure}[ht]
    \begin{center}
       \includegraphics[width=\linewidth]{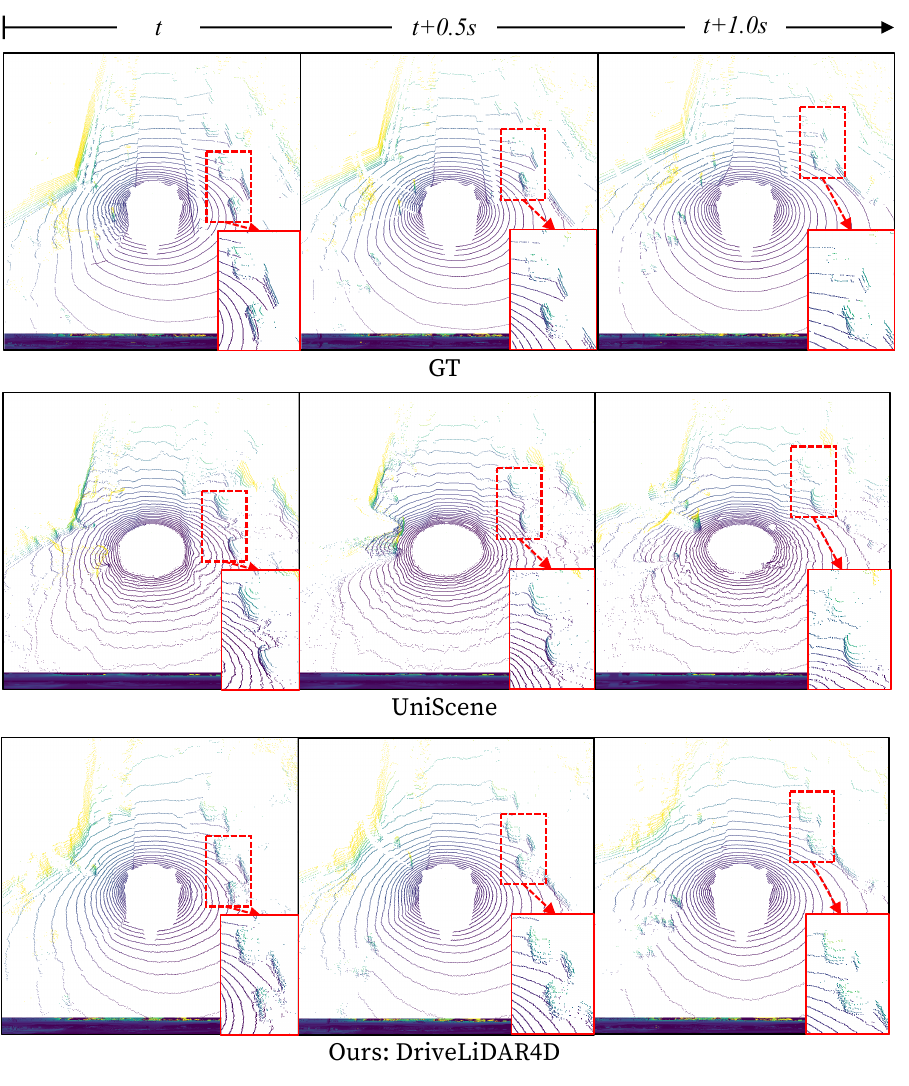}
    \end{center}
       \caption{Sequential LiDAR scenes generated by different methods on the nuScenes val split. Note that \sysname\ generates sequences of 20 frames, of which we present three representative frames for a clear comparison. (Please refer to the supplementary material for more visual comparisons.)}
       \label{fig_seq_generation}
\end{figure}

We compare \sysname\ with the SOTA sequential LiDAR generation methods, LidarDM \cite{zyrianov2024lidardm} and UniScene \cite{li2024uniscene}. UniScene is a joint RGB-LiDAR generation approach, and we evaluate the quality of its generated LiDAR scenes. It is important to note the difference in sequence length capabilities: LidarDM and UniScene generate sequences of 5 and 8 frames respectively, whereas \sysname\ generates 20 frames at a time. The FVD metric is computed using the the first 5 frames of all the generated sequences to ensure a fair comparison.

\cref{tab_seq_con} presents the results of different methods on the nuScenes val split. It can be seen that \sysname\ achieves lower FRD, MMD and JSD scores than LidarDM and UniScene, indicating improved spatial fidelity and diversity in the generated scenes. In addition, \sysname\ attains the lowest FVD \cite{li2024uniscene} score across the comparing methods, indicating superior temporal consistency in the generated sequences .

\cref{fig_seq_generation} illustrates the qualitative comparison of the sequential LiDAR scene generated by different methods on the nuScenes val split. It can be seen that the objects' structures are mostly broken in UniScene's generated scenes, and the objects' locations diverges from the GT locations  (as evident in the zoomed-in views). In contrast, \sysname\ yields better objects structures and accurate locations, benefiting from the rich information provided by multimodal conditioning. Furthermore, \sysname\ exhibits better temporal consistency with coherent vehicle representations across frames, which is attributed to the EST-Conv and EST-Trans components of \modelname.

\subsection{The Impact of Model Configuration}

We study the impact of EST-Conv and EST-Trans of \modelname\ by evaluating the following model configurations: \textit{Config A} refers to the model where we remove EST-Trans in \sysname\ and replace EST-Conv with regular 2D Convolution blocks. \textit{Config B} refers to the model where we remove EST-Trans in \sysname. \textit{Config C} is the \sysname\ model.

By comparing \textit{Config A} and \textit{Config B} in \cref{tab_ablation_multi}, it is evident that the EST-Conv is critical in enhancing the temporal consistency of the generated LiDAR scenes. By capturing and correlating local features across consecutive frames, the EST-Conv module enables the model to generate temporally coherent and high-quality point clouds. Consequently, both FRD and FVD scores are reduced.
Furthermore, \textit{Config C} yields additional FRD and FVD improvements compared to \textit{Config B}, indicating that EST-Trans substantially enhances spatio-temporal consistency in the generated point clouds.

\begin{table}
\centering
\begin{tabular}{lcc|cc} 
\toprule
  & \begin{tabular}[c]{@{}c@{}}EST-Conv\end{tabular} & \begin{tabular}[c]{@{}c@{}}EST-Trans\end{tabular} & FRD$\downarrow$ & FVD$\downarrow$ \\ 
\midrule
\textit{Config A} &                                                       &                                                        & 318.05 & 29.50  \\
\textit{Config B} & \checkmark                                             &                                                       & 311.38  & 17.28 \\
\textit{Config C} & \checkmark                                             & \checkmark                                            & \textbf{292.28}   & \textbf{14.86}\\
\bottomrule

\end{tabular}
\caption{
Sequential generation results of \sysname\ with different module configurations.}
\label{tab_ablation_multi}
\end{table}

\subsection{Real-world 3D Object Detection Evaluation}

\begin{table}[h]
\centering
\scalebox{0.98}{
\begin{tabular}{l|cc} 
\toprule
         & mAP $\uparrow$          & GT Agg.$^1$$\uparrow$                  \\ 
\midrule
GT       &   0.804                   & 100.0\%                      \\
UniScene \cite{li2024uniscene} & 0.078 &   9.7\% \\
LidarDM \cite{zyrianov2024lidardm} & 0.140 &   17.4\% \\
\rowcolor{color_ours}
\sysname\ w.o. object priors & 0.123 &  15.3\%  \\
\rowcolor{color_ours}
\sysname\  &\textbf{0.407} & \textbf{50.6}\%  \\
\bottomrule
\multicolumn{3}{l}{\footnotesize 1: GT Agg. denotes aggrements with mAP of GT.} \\
\end{tabular}
}
\caption{3D object detection results on LiDAR scenes generated by different method on the nuScenes val split. (Please refer to the supplementary material for a visualization of detected objects.)}
\label{tab_object_detection}
\end{table}

The accuracy of object localization and the fidelity of objects' points are essential in real-world autonomous driving scenarios, as these factors significantly impact the performance of perception systems.
\cref{tab_object_detection} presents the 3D object detection results for \textit{cars} on LiDAR scenes generated by different methods on the nuScenes val split. It can be observed that UniScene and LidarDM only achieved a mAP of 0.078 and 0.140, respectively, lagging significantly from the results on the real-world GT data. In comparison, \sysname\ achieves a mAP of 0.407, indicating strongest agreement, 50.6\%, with GT data. The ablated variant, \sysname\ w.o. object priors, yields a mAP of 0.123, showing that object priors plays an important role in enhancing object-level fidelities. This is because object priors provide direct and dense guide on objects' point distributions. 

\section{Conclusion}

In this paper, we introduce \sysname, a novel 4D LiDAR scene generation pipeline that incorporates three multimodal conditions and a equirectangular spatial-temporal noise prediction model, \modelname. The multimodal conditions enable precise scene manipulation, encompassing both foreground objects, background elements and object-level fidelity enhancement. Concurrently, \modelname\ ensures spatial and temporal consistency in the generated sequential LiDAR scenes.
We believe that the realistic 4D LiDAR scenes generated by \sysname\ would contribute significantly to the development and evaluation of real-world autonomous driving systems.


\bigskip

\bibliography{aaai2026}

\end{document}